\documentclass[reprint,twocolumn,amsmath,amssymb,aps,superscriptaddress,prb,eqsecnum]{revtex4-1}
\usepackage{graphicx}
\usepackage{bm}
\usepackage{epsfig}
\usepackage{amssymb}
\usepackage{amsfonts}
\usepackage{amsmath}
\usepackage{color}
\usepackage{xcolor}
\usepackage{mathdesign}
\usepackage{epstopdf}
\usepackage{hyperref}
\usepackage{float}
\usepackage{appendix}
\restylefloat{table}
\usepackage{wasysym}
\usepackage{bibentry}
\usepackage{multirow}
\usepackage[caption=false]{subfig}
\usepackage{braket}
\newcommand{\ba}{\begin{eqnarray}}
\usepackage{manfnt}
\newcommand{\ea}{\end{eqnarray}}
\newcommand{\bd}{\begin{displaymath}}
\renewcommand{\v}[1]{{\bf #1}}
\newcommand{\nn}{\nonumber \\}
\DeclareGraphicsExtensions{.pdf,.png,.jpg}

\begin{document}
\title{Fluctuation-dissipation Type Theorem in Stochastic Linear Learning}
\author{Manhyung Han}
\affiliation{School of Electrical Engineering, Seoul National University, Seoul, Korea}
\email{hanmanhyung@gmail.com}
\author{Jeonghyeok Park}
\affiliation{Cambridge University, Jesus College, Jesus Ln, Cambridge CB5 8BL, UK}
\email{jp868@cam.ac.uk}
\author{Taewoong Lee}
\affiliation{Harvard College, Harvard University, Cambridge, MA 02138, United States}
\email{taewoonglee@college.harvard.edu}
\author{Jung Hoon Han}
\affiliation{Department of Physics, Sungkyunkwan University, Suwon 16419, Korea}
\email{hanjemme@gmail.com}
%
\begin{abstract} The fluctuation-dissipation theorem (FDT) is a simple yet powerful consequence of the first-order differential equation governing the dynamics of systems subject simultaneously to dissipative and stochastic forces. The linear learning dynamics, in which the input vector maps to the output vector by a linear matrix whose elements are the subject of learning, has a stochastic version closely mimicking the Langevin dynamics when a full-batch gradient descent scheme is replaced by that of stochastic gradient descent. We derive a generalized FDT for the stochastic linear learning dynamics and verify its validity among the well-known machine learning data sets such as MNIST, CIFAR-10 and EMNIST. 
\end{abstract}
\maketitle

\section{Introduction}
It is not an uncommon perception among the practitioners of machine learning and of theoretical many-body physics that some ideas of physics, most notably those of equilibrium and nonequilibrium statistical physics, might have significance in the fundamental understanding of the machine learning dynamics. Such sentiment and progress along the direction has continued for some time and still in active pursuit, mostly by the researchers in the machine learning community~\cite{welling11,fox15,mandt16,chaudhari18,baldassi19,ganguli13,ganguli19,ganguli20,yaida18}. The belief in the statistical-physics foundation of the machine learning will be strengthened obviously by more examples of ideas originating from statistical physics and then manifesting themselves in the machine learning. Here we establish one such connection, relating a fundamental theorem in near-equilibrium statistical physics~\cite{ao04,KAT05,ao06,kwon11a,kwon11} to the theory of learning dynamics~\cite{welling11,fox15,mandt16,yaida18,ganguli13,ganguli19}, in particular where the learning process is {\it linear} and described by a stochastic equation similar to what governs the Ornstein-Uhlenbeck processes~\cite{risken96}. The theorem in question is the fluctuation-dissipation theorem (FDT). 

The FDT in a strict sense refers to specific relations that hold between correlation functions and response functions of physical systems under equilibrium~\cite{risken96}. Here we use the term in a more relaxed sense, referring to mathematical identities among the observable quantities under the stationary state condition. The difference between the equilibrium and the stationary state is revealed by the existence of an anti-symmetric matrix $\v Q$~\cite{ao04,KAT05,ao06,kwon11a,kwon11}, which will be defined shortly. The FDT is illustrated most simply in the Langevin dynamics of a single particle subject simultaneously to dissipative and stochastic forces 
\ba \dot{x} = -\gamma x + f(t) \ea
where, in the context of Newtonian motion, $x$ represents the velocity of a particle in one dimension, $-\gamma x$ is the resistive force, and $f(t)$ is the random force coming from the environment. On integrating the first-order differential eqution we obtain the formally exact solution $x(t) = e^{-\gamma t} [ x(0) + \int_0^t e^{\gamma t'} f(t')  ]$ which, in the long-time limit ($t\rightarrow \infty$) yields the average 
\ba 
\langle x^2 \rangle = 2D e^{-2\gamma t}  \int_0^t dt' e^{2\gamma t'}  = D/\gamma \ea
assuming the white-noise correlation
$\langle f(t) f(t') \rangle = 2D \delta (t - t' )$. The competing tendencies of the dissipation ($\gamma$) and fluctuation ($D$) finds balance through the identity. 

Multi-dimensional generalization of the Langevin dynamics finds expression in 
\ba \dot{\v x} = -{\bf \Gamma} {\v x}  + {\v f} ( t)  \label{eq:1.1} \ea
with $n$-dimensional variables $\v x = (x_1 , \cdots x_n )$, the $n\times n$ dissipation matrix $\bf \Gamma$, and the $n$-dimensional stochastic force vector $\v f$ obeying the zero mean $\langle \v f \rangle =0$ and the variance $\langle \v f ( t) \v f^T (t' ) \rangle = 2 \v D \delta (t-t')$, in terms of the $n\times n$ diffusion matrix $\v D$. From the exact solution $\v x(t) = e^{-\v \Gamma t} [ \v x (0) + \int_{0}^{t} e^{\v \Gamma t'} \v f (t') dt' ]$ we derive the long-time correlation average 
\ba
\bm \Sigma(t) & = & \langle \v x(t) \v x^T(t) \rangle \nn
%
%
& = & 2 \int_{0}^{t} dt' e^{\v \Gamma (t'-t)} \v D e^{\v \Gamma^T(t'-t)}
\ea
and the following identity for $\bm \Sigma = \bm \Sigma (t \rightarrow \infty)$: 
\ba
\v \Gamma \bm \Sigma + \bm \Sigma \v \Gamma^T = 2\v D .  \label{eq:2.13} 
\ea
This identity relates the diffusion matrix $\v D$ with the dissipation matrix $\bf  \Gamma$ through the correlation matrix $\bm \Sigma$ in the stationary-state, for the Ornstein-Uhlenbeck processes with constant $\bf \Gamma$ and $\v D$~\cite{ao04,KAT05}. Extensions and applications of the theorem both in physical systems and machine learning have since appeared~\cite{kwon11,fox15,mandt16}. Thanks to the identity, one can write the matrix $\bf \Gamma \bm \Sigma$ as the sum of the symmetric ($\v D$) and anti-symmetric ($\v Q$) matrix:
\ba \bm \Gamma \bm \Sigma = \v D + \v Q  . \label{eq:decomposition} \ea
It was pointed out in Ref. \onlinecite{kwon11a} that $\v Q =0$ implies the detailed balance, otherwise one should allow the possibility $\v Q \neq 0$ in the decomposition, Eq. (\ref{eq:decomposition}).

In Sec. \ref{sec:FDT-for-W}, we derive an analogous mathematical identity for the stochastic linear learning dynamics. This is then verified, in Sec. \ref{sec:experiments}, through numerical experiments on several well-known machine learning datasets. Implications of our work are discussed in Sec. \ref{sec:discussion}.

\section{FDT in Learning Dynamics} 
\label{sec:FDT-for-W}

In the learning dynamics one is confronted with a collection of input vectors $\v x_\alpha$ (e.g. pixels in a jpg file re-formatted as a one-dimensional vector) and output vectors $\v y_\alpha$ (e.g. classification of the picture as an image of a cat or a dog), where $1 \le \alpha \le N$ runs over the entire dataset called the {\it batch}. In the linear learning dynamics one is interested in finding the matrix $\v W$ that minimizes the error
\ba E & = & \frac{1}{2N} \sum_{\alpha=1}^N ({\bf y}_\alpha - \v W \v x_\alpha )^T ({\bf y}_\alpha - \v W \v x_\alpha ) \nn
& \equiv &  \frac{1}{2} {\rm Tr} [ {\bf \Sigma}_{xx} {\bf W}^T {\bf W}  - {\bf W}^T {\bf \Sigma}_{yx} - {\bf \Sigma}^T_{yx} {\bf W}  ] . \label{eq:error-function}
\ea
The two correlation functions appearing in the second line are 
\ba {\bf \Sigma}_{xx} = \frac{1}{N} \sum_{\alpha=1}^N {\bf x}_\alpha {\bf x}_\alpha^T , ~~ {\bf \Sigma}_{yx} = \frac{1}{N} \sum_{\alpha=1}^N  {\bf y}_\alpha {\bf x}_\alpha^T. \label{eq:Sigma-xx-definition}\ea
The gradient descent (GD) method of finding the optimal $\v W$ results in the first-order differential equation for $\v W$~\cite{ganguli13,ganguli19}:
\ba \frac{d \bf W }{dt} =  - \frac{\delta E}{\delta \v W} = - {\bf W} {\bf \Sigma}_{xx} + { \bf \Sigma}_{yx} . \label{eq:1.10} \ea
The full solution is given by ${\bf W}(t) = {\bf W}(0) e^{-{\bf \Sigma}_{xx} t} + \v W_0 (1 - e^{-{\bf \Sigma}_{xx} t} )$ where $\v W_0 = {\bf \Sigma}_{yx} {\bf \Sigma}_{xx}^{-1}$ offers the equilibrium solution. 

An interesting connection to the Langevin dynamics and FDT arises when we treat $\bm \Sigma_{xx}$ and $\bm \Sigma_{yx}$ in the dynamics of Eq.  (\ref{eq:1.10}) as a mini-batch (not a full-batch) average. At each stage of $\v W$-evolution one picks a different, randomly chosen mini-batch to compute the average $\bm \Sigma_{xx} (t) = N_m^{-1} \sum_{\alpha \in B(t)} {\bf x}_\alpha {\bf x}_\alpha^T$ and ${\bf \Sigma}_{yx} (t) =  N_m^{-1} \sum_{\alpha \in B(t) }  {\bf y}_\alpha {\bf x}_\alpha^T$, where $N_m$ is the mini-batch size and $B(t)$ is the particular mini-batch chosen at the time $t$. The $\v W$-dynamics according to the stochastic gradient descent (SGD) scheme becomes 
\ba \frac{d \v W}{dt} = - \v W \bm \Sigma_{xx}(t) + \bm \Sigma_{yx} (t) . \label{eq:stochastic-W-dynamics} \ea
Phrased in the language of Langevin dynamics, both the dissipative ($\bm \Sigma_{xx} (t)$) and the stochastic ($\bm \Sigma_{yx} (t)$) forces are time-dependent. We can re-write the variables in the equation explicitly as the sum of the stationary (time-independent) and the fluctuating (time-dependent) parts,
\ba \v W (t) & \rightarrow & \v W_0 + \v W (t), \nn
\bm \Sigma_{xx} (t) & \rightarrow & \bm \Sigma_{xx} + \bm \Sigma_{xx} (t), \nn
\bm \Sigma_{yx} (t) & \rightarrow &\bm \Sigma_{yx} + \bm \Sigma_{yx} (t),  \label{eq:re-definition} \ea
and work with the equation
\ba
\frac{d\v W }{dt}\! =\!  -\v W ( \bm \Sigma_{xx}  \!+\! \bm \Sigma_{xx} (t) ) \!+\! \bm \Sigma_{yx} (t) \!-\! \v W_0 \bm \Sigma_{xx}  (t) . \nonumber \\ \label{eq:modified-stochastic-DE}  \ea
Although the exact solution to this equation can be found in the form of Wiener integral (see Appendix \ref{appendix:A}), we will here assume a simplified situation where $\bm \Sigma_{xx} (t) =0$ on the right-hand side of the equation. Relaxing the assumption will not change the overall conclusion as long as $\bm \Sigma_{xx} (t)$ is small - see Appendix \ref{appendix:A}. The stochastic learning dynamics is now reduced to an Ornstein-Uhlenbeck process~\cite{risken96} and allows a simple solution
\ba \v W(t) = \left[ \v W (0) + \int_0^t \bm \Sigma_{yx} (t') e^{\bm \Sigma_{xx} t' } \right] e^{-\bm \Sigma_{xx} t } . \label{eq:approximate-W} \ea

We can write down the long-time correlation matrix
\ba
\bm \Sigma_{WW} (t) & = & \langle \v W^T(t) \v W(t) \rangle \nn
& = & \int_{0}^{t} dt' \int_{0}^{t} dt'' e^{\bm \Sigma_{xx} (t'-t)} \langle \bm \Sigma_{yx}^T(t')\bm \Sigma_{yx} (t'') \rangle e^{\bm \Sigma_{xx} (t''-t)} \nn
& = & \int_{0}^{t} dt' e^{\bm \Sigma_{xx} (t'-t)} 2\v D e^{\bm \Sigma_{xx} (t'-t)}
\ea
assuming $ \langle \bm \Sigma_{yx}^T(t')\bm \Sigma_{yx} (t'') \rangle = 2\v D\delta(t'-t'')$. From this follows the  identity 
\ba
\bm \Sigma_{xx} \Sigma_{WW} + \bm \Sigma_{WW} \bm \Sigma_{xx}   = 2\v D \label{eq:FDT-for-W}
\ea
for $\bm \Sigma_{WW}  \equiv \bm \Sigma_{WW} (t \rightarrow \infty )$. This is the FDT type identity in the stochastic linear learning dynamics and our central result (a more refined form of FDT exists - see Appendix \ref{appendix:B}). In the expression (\ref{eq:FDT-for-W}), $\bm \Sigma_{xx}$ is the full-batch correlation matrix given in Eq. (\ref{eq:Sigma-xx-definition}). Restoring the original definition, we have 
\ba 
\langle  ( \bm \Sigma_{yx} (t)  \!-\! \bm \Sigma_{yx} )^T ( \bm \Sigma_{yx} (t') \!-\! \bm \Sigma_{yx} ) \rangle & = &  2\v D \delta(t -t') \nn
\langle  [ \v W (t) - \v W_0 ]^T  [\v W (t) - \v W_0 ]  \rangle & = & \bm \Sigma_{WW} . 
\label{eq:definitions} 
\ea

The full-batch input-input correlation matrix $\bm \Sigma_{xx}$ provides a sort of dissipative force while (fluctuating part of) the input-output correlation function plays the stochastic force in the learning dynamics, according to Eq. (\ref{eq:modified-stochastic-DE}). The correlator of the learning matrix, i.e. $\bm \Sigma_{WW}$, is obtained as the balance between the two tendencies. 

%
%

\section{Numerical Experiments}
\label{sec:experiments} 

For sufficiently small time $t=h$ we can solve the stochastic equation (\ref{eq:stochastic-W-dynamics})  approximately
\ba
\v W (h) &\approx & \left[ \v W (0) + \int_{0}^{h} \bm \Sigma_{yx} (t')e^{\bm \Sigma_{xx} (0) t'} dt' \right]e^{-\bm \Sigma_{xx} (0) h} \nn
& \approx & \v W (0) [1 - \bm \Sigma_{xx} (0) h ] + \int_{0}^{h} \bm \Sigma_{yx} (t') dt' . \label{eq:W-h} 
\ea
We can further divide up the interval $t \in [0, h]$ into $M$ equal segments, each of width $\varepsilon \equiv h/M$, and use the discrete formula
$\Sigma_{xx} (0) \rightarrow M^{-1} \sum_{i = 1}^M \Sigma_{xx} (i \cdot \Delta)$ and 
$ \int_{0}^{h} \bm \Sigma_{yx} (t') dt' \rightarrow \varepsilon \sum_{i=1}^M \Sigma_{yx} ( i \cdot \Delta )$. In the end, Eq. (\ref{eq:W-h}) turns into a recursive formula
\ba \v W^{(n+1)} = \v W^{(n)} [ 1- \varepsilon \bm \Sigma_{xx}^{(n)} ] + \varepsilon \bm \Sigma_{yx}^{(n)}
\label{eq:W-n} \ea
where $\bm \Sigma_{xx}^{(n)}$ and $\bm \Sigma_{yx}^{(n)}$ are averages over the mini-batch of size $M N_m$. At sufficiently large $n$, $\v W^{(n)}$ executes a steady-state fluctuation around the minimum $\v W_0$. 

\begin{figure*}[ht]
\centering
\includegraphics[width=0.8\textwidth]{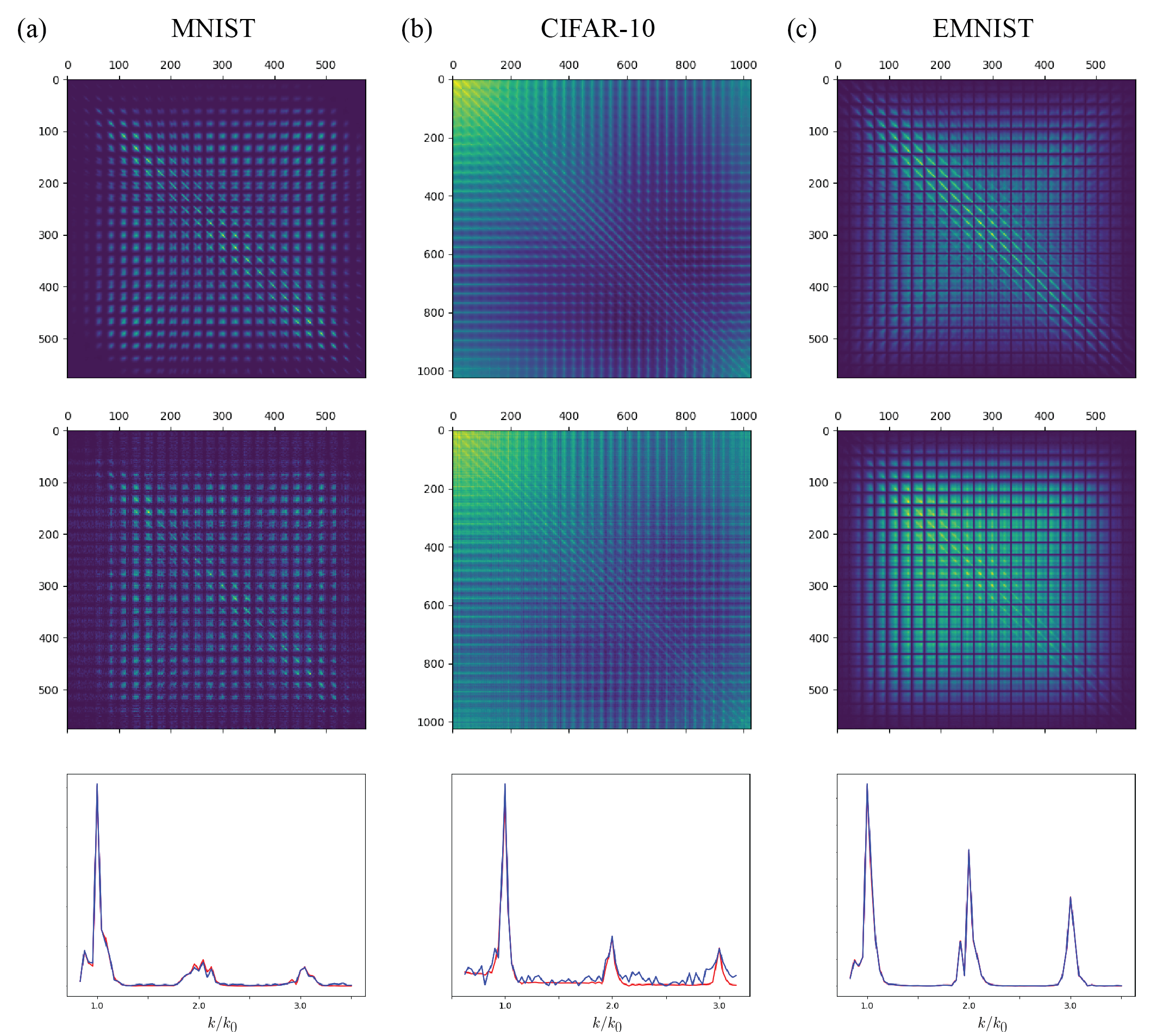}
\caption{Fluctuation analysis for (a) MNIST (b) CIFAR-10 and (c) EMNIST datasets. (top) Plots of $\v D$ obtained from each dataset. (middle) Plots of $\bm \Sigma_{xx} \bm \Sigma_{WW} + \bm \Sigma_{WW} \bm \Sigma_{xx}$. (bottom) Normalized Fourier components for $\v D$ (red) and $\bm \Sigma_{xx} \bm \Sigma_{WW} + \bm \Sigma_{WW} \bm \Sigma_{xx}$ (blue) plotted along $\v k = (k_x , 0)$ with $k_0 = 2\pi/a$.} 
\label{fig:MNIST-CIFAR}
\end{figure*}

To test out the validity of the FDT in stochastic linear learning derived in Eq. (\ref{eq:FDT-for-W}), we employ three representative datasets: MNIST, CIFAR-10 and EMNIST Letters (abbreviated as EMNIST from here on)~\cite{cohen17}. MNIST and CIFAR-10 consist of ten different objectives or output vectors $\v y^\alpha$, represented by one-hot vectors $(1,0, \cdots, 0)$ through $(0, \cdots, 0, 1)$. Twenty-six alphabets are represented by as many output vectors in the case of EMNIST. The pixel sizes are $28\times28$ for both MNIST and EMNIST, and $32\times32$ for CIFAR-10. Updating $\v W (t)$ according to the SGD algorithm outlined in Eqs. 
(\ref{eq:W-h}) and (\ref{eq:W-n}), we found good convergence to the error-minimizing value $\v W_0 = \bm \Sigma_{xy} \bm \Sigma_{xx}^{-1}$ by measuring the inner product of the $\v W(t)$ and $\v W_0$ divided by their norms approaching unity: $\cos \theta(t) = \v W (t) \cdot \v W_0 / \| \v W (t) \| \| \v W_0 \|$. The inner product of two matrices is defined by taking a product of the matrix elements sharing the same $(ij)$ index and making a sum over all $(ij)$'s.

Once the steady state is reached, e.g. $\cos \theta \gtrsim 0.999$, we begin analyzing the small fluctuations by calculating the two correlators in Eq. (\ref{eq:definitions}) by taking averages $\langle \cdots \rangle$ over several tens of thousands of $\v W^{(n)}$'s and $\bm \Sigma_{yx}^{(n)}$'s. To deduce the diffusion matrix $\v D$ in Eq. (\ref{eq:definitions}) we take the equal-time correlator $t=t'$ and compute the average of $\bm \Sigma_{yx}^{(n)}$. This gives the $\v D$ matrix up to an overall constant. In the end, a good proportionality between $\bm \Sigma_{xx} \bm \Sigma_{WW} + \bm \Sigma_{WW} \bm \Sigma_{xx}$ and $\v D$ is found as shown in Fig. \ref{fig:MNIST-CIFAR} for all the datasets tested. It turns out the correlators exhibit a highly periodic structure with period $a$ coming from the $a\times a$ pixel size of each dataset. (The original $a=28$ dimension of the MNIST and EMNIST was chopped at the boundary to $a=24$. Otherwise it was difficult to get the full-batch inverse $\bm \Sigma_{xx}^{-1}$.) 

Due to the highly periodic structure of the real-space images of $\bm \Sigma_{xx} \bm \Sigma_{WW} + \bm \Sigma_{WW} \bm \Sigma_{xx}$ and $\v D$, only a handful of Fourier peaks at $\v k= (k_x , k_y)$ given by multiples of $2\pi/a$ were significant. Figure \ref{fig:MNIST-CIFAR} shows the Fourier components along $\v k = (k_x , 0)$ normalized by the value at $\v k = (0,0)$. 
The near-perfect match in the Fourier analysis of both  $\bm \Sigma_{xx} \bm \Sigma_{WW} + \bm \Sigma_{WW} \bm \Sigma_{xx}$ and $\v D$ is not {\it a priori} obvious, and must be attributed to the FDT theorem at work in the stochastic linear learning dynamics. 

\section{Discussion}
\label{sec:discussion}

Our work addresses a FDT type relation in the stochastic linear learning dynamics. The relation derived in Eq. (\ref{eq:FDT-for-W}) is found to hold quite well for a number of machine learning datasets. The analogy to the Langevin dynamics naturally gives rise to an interpretation of the input covariance matrix $\bm \Sigma_{xx}$ as the effective friction, and the input-output variance $\bm \Sigma_{yx}$ as the effective stochastic force in the learning dynamics. 

We have made several attempts to go beyond the simple stochastic linear learning scheme. For one, we tried placing a CNN layer before the neural network layer $\v W$. As shown in Appendix \ref{appendix:C}, this formulation naturally leads to FDT in terms of the CNN-filtered input data sets $\v X^\alpha = \v C \otimes \v x^\alpha$, where $\otimes$ represents the CNN operation. The FDT holds with respect to the renormalized datasets $\v X^\alpha$. In another attempt, we tried introducing non-linearity explicitly by using an alternative error function $E = (2N)^{-1} \sum_{\alpha =1}^N \sum_{i=1}^n (y^\alpha_i - z^\alpha_i )^2$ with the sigmoid function $z^\alpha_i = [ e^{-\sum_{j=1}^n W_{ij} x_j^\alpha} + 1 ]^{-1}$ parameterized by the learning matrix $\v W$. Such formulation leads to the dynamics $d \v W /dt$ that is, unfortunately, highly non-linear and defies further analytical treatment. 

The FDT type relation in the stochastic learning was noticed some years earlier by Yaida~\cite{yaida18}. His derivation of the so-called FDT relation avoids any use of an explicit error function and relies solely on the stationary property of observables after the learning process has saturated. It is a powerful formulation in the sense that the relations apply to an arbitrary learning architecture with non-linearities. On the other hand, by avoiding the stochastic differential equation formulation, the connection that his relations have with the FDT in statistical physics becomes somewhat vague. More seriously, when our error function is used to work out his formulas, the outcome does not match our FDT formula derived in Eq. (\ref{eq:FDT-for-W}). This leads us to suspect that there may be multiple FDT type theorems governing the stationary states of learning, with both our formula and his addressing different facets. 

We have investigated whether, writing $\bm \Sigma_{xx} \bm \Sigma_{WW}$ in Eq. (\ref{eq:FDT-for-W}) as the sum $\bm \Sigma_{xx} \bm \Sigma_{WW} = \v D + \v Q$, there will be a significant contribution of the anti-symmetric matrix $\v Q$. A crude measure of the significance of $\v Q$ relative to $\v D$ is the maximum value of the matrix elements in $\v Q$ divided by that of $\v D$. The results are 0.12, 0.096, 0.045 for MNIST, CIFAR-10, and EMNIST, respectively, suggesting that the anti-symmetric components are probably very small and insignificant. 

\acknowledgments 
The Python code used in the numerical experiment can be found at https://github.com/lemonseed117/FDT-Stochastic.git  J. H. H. acknowledges fruitful discussion with and input on the manuscript from Ping Ao, J. H. Jo, S. B. Lim, J. D. Noh, Vinit Singh, and Hayong Yun.

\appendix 
\section{Full Wiener Integral}
\label{appendix:A} 
The fluctuation-dissipation theorem  (\ref{eq:FDT-for-W}) for stochastic linear learning dynamics was obtained assuming time-independent $\bm \Sigma_{xx}$ and time-dependent $\bm \Sigma_{yx}(t)$. It means a full-batch $\Sigma_{xx}$ and a mini-batch $\bm \Sigma_{yx}(t)$ are assumed in the derivation. On the other hand, the numerical integration of Eq. (\ref{eq:stochastic-W-dynamics}) or (\ref{eq:modified-stochastic-DE}) was done in our experiment using both mini-batch $\Sigma_{xx} (t)$ and $\Sigma_{yx} (t)$. In spite of the difference, the FDT seems to hold quite well numerically. We re-examine the full stochastic equation for linear learning dynamics written in Eq. (\ref{eq:modified-stochastic-DE}),

\ba
\frac{d\v W }{dt} &=&  -\v W ( \bm \Sigma_{xx}  \!+\! \bm \Sigma_{xx} (t) ) \!+\! \bm \Sigma_{yx} (t) \!-\! \v W_0 \bm \Sigma_{xx}  (t) \nn
& = &  -\v W ( \v A  + \v a (t) ) + \v b(t) . \ea
Notations have been simplified in the second line. The full solution to this can be found using the Wiener path integral formulation, familiarly known in physics as the Feynman path integral. 

First, one makes a decomposition $\v W  \rightarrow \v W e^{-\v A t}$, and derive the equation in terms of the new $\v W$:

\ba
{d\v W \over dt} & = &  -\v W e^{-\v A t} \v a (t) e^{\v A t}  + \v b(t) e^{\v A t}  \nn
& = & - \v W \v a_I (t) + \v b_I (t) . \ea
The subscript $I$ is meant to indicate the ``interaction picture'' representation of the learning dynamics following a similar jargon in quantum mechanics. With both $\v a_I (t)$ and $\v b_I (t)$ being time-dependent, one can find the solution in the path integral form

\begin{widetext}
\ba \v W (t) & = &  \left( \v W (0) + \int_0^t dt' \v b_I (t' ) P_{t' \rightarrow 0} \exp [ \int_0^{t'} dt'' \v a_I (t'' ) ] \right) P_{0 \rightarrow t} \exp [- \int_0^t \v a_I (t' ) dt' ] \nn
& = & \v W(0) P_{0 \rightarrow t} \exp [- \int_0^t \v a_I (t' ) dt' ] + \int_0^t dt' \v b_I (t' ) P_{t' \rightarrow t} \exp [-\int_{t'}^t \v a_I (t'' ) dt'' ] .  \label{eq:Wiener-integral} \ea
\end{widetext}
The symbol $P_{0 \rightarrow t}$ means that operator (matrix) defined at time $t' = 0$ is to be written at the far left, and the one at time $t' = t$ at the far right. The symbol $P_{t' \rightarrow t}$ means that $t'' = t'$ operator appears on the far left, and $t''= t$ operator at the far right. The usual composition rule of path integrals gives the second expression of the second line. 

The first term $\sim \v W (0)$ can be ignored because the long-time result should not depend on the initial condition. Furthermore, we are interested in terms that are only first-order in the fluctuation. Under these assumptions we can write the result in Eq. (\ref{eq:Wiener-integral}) approximately 
\ba \v W(t) - \v W_0  \approx \int_0^t dt' \v b (t') e^{\v A (t' - t)  } . \label{eq:Wiener-W} \ea
The full definition of the $\v W$ matrix is restored in the above. Note that this is exactly the same expression obtained earlier in Eq. (\ref{eq:approximate-W}), without the initial $\v W(0)$. Hence the FDT derived earlier is valid to the leading order in the fluctuation.

\section{Refinement of the FDT Theorem}
\label{appendix:B}
It is possible to define a a more general kind of diffusion matrix than the one presented in Eq. (\ref{eq:definitions}),
\ba 
& \langle  ( \bm \Sigma_{yx} (t)  \!-\! \bm \Sigma_{yx} )_{i\alpha} ( \bm \Sigma_{yx} (t') \!-\! \bm \Sigma_{yx} )_{j\beta} \rangle \nn
& =   2 D_{i\alpha, j\beta}  \delta(t -t') \ea
that does not involve the summation over the output indices $i,j$.
%
%
A similar generalization for the $W$-correlation matrix gives
\ba
& & \langle ( W - W_0 )_{ij} (W- W_0 )_{kl} \rangle = \Sigma_{ij,kl}  \nn
%
%
& = & \int_{0}^{t} dt' 2D_{i\alpha,k\beta} [e^{\v A (t'-t)}]_{\alpha j} [e^{\v A (t'-t)}]_{\beta l}, 
\ea
where the result from Eq. (\ref{eq:Wiener-W}) is used to reach the second line. 

If we fix the two output indices $i$ and $k$ in the above relation, then one can re-write it in the following fashion: 
\ba
(\Sigma_{i,k})_{jl} & = & \int_{0}^{t} dt' [e^{\v A^T(t'-t)}]_{j \alpha} (2D_{i,k})_{\alpha\beta} [e^{\v A(t'-t)}]_{\beta l} \nn
& = & \int_{0}^{t} dt' [e^{\v A^T(t'-t)} 2D_{i,k} e^{\v A(t'-t)}]_{jl} .
\ea
In other words, for a given pair of output indices $(i,k)$, we have a matrix relation
\ba
\bm \Sigma_{i,k}  = \int_{0}^{t} dt' [e^{\v A^T(t'-t)} 2\v D_{i,k} e^{\v A(t'-t)}], 
\ea
subject to the same kind of identify as before: 
\ba
\v A^T \bm \Sigma_{i,k} + \bm \Sigma_{i,k}\v A & = & \int_{0}^{t} dt' {d \over dt'} [e^{\v A^T(t'-t)} 2\v D_{i,k} e^{\v A(t'-t)}] \nn
& = & 2\v D_{i,k} . 
\ea
In conclusion, the FDT holds irrespective of the choice of output indices $(i,k)$. 

\section{Stochastic Linear Learning \\ with CNN layer}
\label{appendix:C}

Adding a CNN layer before the $\v W$ layer and optimizing the error function with respect to both $\v W$ and the CNN filter matrix turns out to be within the mathematically tractable scope. The CNN layer transform the input vector $\v x^\alpha$ into a modified input vector $\v X^\alpha = \v C \otimes \v x^\alpha$ according to the recipe,
\ba X^\alpha_{\bm j} = \sum_{i_x , i_y =1}^c C_{\bm i} x^\alpha_{\bm i  + \bm j  - \v 1} .  \ea
We switch to a two-dimensional vector notation of the indices, $\bm i = (i_x , i_y )$, $\bm j = (j_x , j_y )$, and write $\v 1 = (1,1)$. The convolution operator $\v C$ is $c\times c$ dimensional matrix. The error function is the same as before, Eq. (\ref{eq:error-function}) with $\v X^\alpha$ taking the place of $\v x^\alpha$:
\ba E & = & \frac{1}{2N} \sum_{\alpha=1}^N ({\bf y}_\alpha - \v W \v X_\alpha )^T ({\bf y}_\alpha - \v W \v X_\alpha ) . 
\ea
Minimization of the error must take place with respect to both $\v W$ and $\v C$. 

Taking the derivative of the error $E$ with respect to an element of the convolution matrix $C_{\bm i}$ can be done by using the chain rule,
\ba
\frac{\partial E}{\partial C_{\bm i}} &=& \sum_{\alpha, \bm j} \frac{\partial E}{\partial X_{\bm j}^{\alpha}} \frac{\partial  X_{\bm j }^{\alpha}}{\partial C_{\bm i}} \nn
&=& \sum_{\alpha, \bm j} \frac{\partial E}{\partial X_{\bm j}^{\alpha}} x^\alpha_{\bm j + \bm i  - \v 1} . \label{eq:dE-dC}
\ea
Supplemented by $\delta E / \delta \v X^\alpha = N^{-1} \v W^T (\v W \v X^\alpha - {\bf y}^\alpha)$, we arrive at
\ba \frac{d C_{\bm i} }{dt} =\frac{1}{N} \sum_{\alpha, \bm j} ( \v W^T {\bf y}^\alpha - \v W^T \v W \v X^\alpha )_{\bm j}  x^\alpha_{\bm j + \bm i - \v 1} .   \ea 
%
The first term on the r.h.s. becomes
\ba \frac{1}{N} \sum_{\alpha, \bm j, \bm k } W_{\bm k , \bm j} y^\alpha_{\bm k}  x^\alpha_{\bm j + \bm i - \v 1}  = \sum_{\bm k , \bm j} W_{\bm k, \bm j} \Sigma^{yx}_{\bm k, \bm j + \bm i - \v 1} . \ea
This is a summation over the output index $\bm k$, and a convolution with respect to the input indices. The surviving index is $\bm i$, which covers the elements of the filter matrix $\v C$. We have $\v 1 \le \bm j \le \bm L - \bm c + \v 1 $, $\v 1 \le \bm i \le \bm c$, and $\v 1 \le \bm j + \bm i - \v 1 \le \bm L$, which keeps track of the range of indices in a correct manner. For the second term on the r.h.s. we get
\ba 
&& \frac{1}{N} \sum_{\alpha, \bm j} ( \v W^T \v W \v X^\alpha )_{\bm j}  x^\alpha_{\bm j + \bm i - \v 1} \nn
&=& \frac{1}{N} \sum_{\alpha, \bm j, \bm k } ( \v W^T \v W  )_{\bm j, \bm k} X^\alpha_{\bm k}  x^\alpha_{\bm j + \bm i - \v 1} \nn
&=& \frac{1}{N} \sum_{\alpha, \bm j, \bm k , \bm l  } ( \v W^T \v W  )_{\bm j, \bm k} C_{\bm l} x^\alpha_{\bm l + \bm k - \v 1 }  x^\alpha_{\bm j + \bm i - \v 1} \nn
%
%
&=& \sum_{\bm l  } C_{\bm l} \left[ \sum_{\bm j, \bm k} ( \v W^T \v W  )_{\bm j, \bm k}  (\Sigma^{xx} )_{\bm l + \bm k - \v 1 ,  \bm j + \bm i - \v 1} \right] . \ea
We can define two new quantities 
\ba P_{\bm i, \bm l } & \equiv & 
\sum_{\bm j, \bm k} ( \v W^T \v W  )_{\bm j, \bm k}  (\Sigma^{xx} )_{\bm l + \bm k - \v 1 ,  \bm j + \bm i - \v 1} = P_{\bm l, \bm i} \nn
Q_{\bm i} & \equiv & \sum_{\bm j, \bm k} W_{\bm k, \bm j} \Sigma^{yx}_{\bm k, \bm j + \bm i - \v 1}\label{eq:P-and-Q} \ea 
to simplify the equation
\ba \frac{d C_{\bm i}}{dt} = Q_{\bm i} - \sum_{\bm l} P_{\bm i, \bm l} C_{\bm l} . \label{eq:dC-dt} \ea
The $\v W$ matrix appears in various places in the definition of $\v P$ and $\v Q$, and can be obtained from
\ba \frac{d \v W}{dt } = - \v W \bm \Sigma_{XX} + \bm \Sigma_{yX} \label{eq:dW-dt} \ea
where 
\ba \bm \Sigma^{XX}_{\bm i , \bm j} &=& \sum_{\bm k, \bm l} C_{\bm k} C_{\bm l} \Sigma^{xx}_{\bm k + \bm i - \v 1 , \bm l + \bm j - \v 1 } \nn
\bm \Sigma^{yX}_{\bm i, \bm j} & = & \sum_{\bm k} C_{\bm k}  \bm \Sigma^{yx}_{\bm i, \bm j + \bm k - \v 1}  .  \ea
The two equations (\ref{eq:dC-dt}) and (\ref{eq:dW-dt}) can be solved simultaneously by GSD. At equilibrium we have $\v C= \v P^{-1} \v Q$ but this formula is a bit misleading as the filter $\v C$ enters implicitly in both $\v P$ and $\v Q$ as well. 

%
%

\bibliography{LL}

\begin{thebibliography}{16}%
\makeatletter
\providecommand \@ifxundefined [1]{%
 \@ifx{#1\undefined}
}%
\providecommand \@ifnum [1]{%
 \ifnum #1\expandafter \@firstoftwo
 \else \expandafter \@secondoftwo
 \fi
}%
\providecommand \@ifx [1]{%
 \ifx #1\expandafter \@firstoftwo
 \else \expandafter \@secondoftwo
 \fi
}%
\providecommand \natexlab [1]{#1}%
\providecommand \enquote  [1]{``#1''}%
\providecommand \bibnamefont  [1]{#1}%
\providecommand \bibfnamefont [1]{#1}%
\providecommand \citenamefont [1]{#1}%
\providecommand \href@noop [0]{\@secondoftwo}%
\providecommand \href [0]{\begingroup \@sanitize@url \@href}%
\providecommand \@href[1]{\@@startlink{#1}\@@href}%
\providecommand \@@href[1]{\endgroup#1\@@endlink}%
\providecommand \@sanitize@url [0]{\catcode `\\12\catcode `\$12\catcode
  `\&12\catcode `\#12\catcode `\^12\catcode `\_12\catcode `\%12\relax}%
\providecommand \@@startlink[1]{}%
\providecommand \@@endlink[0]{}%
\providecommand \url  [0]{\begingroup\@sanitize@url \@url }%
\providecommand \@url [1]{\endgroup\@href {#1}{\urlprefix }}%
\providecommand \urlprefix  [0]{URL }%
\providecommand \Eprint [0]{\href }%
\providecommand \doibase [0]{http://dx.doi.org/}%
\providecommand \selectlanguage [0]{\@gobble}%
\providecommand \bibinfo  [0]{\@secondoftwo}%
\providecommand \bibfield  [0]{\@secondoftwo}%
\providecommand \translation [1]{[#1]}%
\providecommand \BibitemOpen [0]{}%
\providecommand \bibitemStop [0]{}%
\providecommand \bibitemNoStop [0]{.\EOS\space}%
\providecommand \EOS [0]{\spacefactor3000\relax}%
\providecommand \BibitemShut  [1]{\csname bibitem#1\endcsname}%
\let\auto@bib@innerbib\@empty
\bibitem [{\citenamefont {Welling}\ and\ \citenamefont
  {Teh}(2011)}]{welling11}%
  \BibitemOpen
  \bibfield  {author} {\bibinfo {author} {\bibfnamefont {M.}~\bibnamefont
  {Welling}}\ and\ \bibinfo {author} {\bibfnamefont {Y.~W.}\ \bibnamefont
  {Teh}},\ }in\ \href@noop {} {\emph {\bibinfo {booktitle} {Proceedings of the
  28th International Conference on International Conference on Machine
  Learning}}},\ \bibinfo {series and number} {ICML'11}\ (\bibinfo  {publisher}
  {Omnipress},\ \bibinfo {address} {Madison, WI, USA},\ \bibinfo {year}
  {2011})\ pp.\ \bibinfo {pages} {681--688}\BibitemShut {NoStop}%
\bibitem [{\citenamefont {Ma}\ \emph {et~al.}(2015)\citenamefont {Ma},
  \citenamefont {Chen},\ and\ \citenamefont {Fox}}]{fox15}%
  \BibitemOpen
  \bibfield  {author} {\bibinfo {author} {\bibfnamefont {Y.-A.}\ \bibnamefont
  {Ma}}, \bibinfo {author} {\bibfnamefont {T.}~\bibnamefont {Chen}}, \ and\
  \bibinfo {author} {\bibfnamefont {E.~B.}\ \bibnamefont {Fox}},\ }in\
  \href@noop {} {\emph {\bibinfo {booktitle} {Proceedings of the 28th
  International Conference on Neural Information Processing Systems - Volume
  2}}},\ \bibinfo {series and number} {NIPS'15}\ (\bibinfo  {publisher} {MIT
  Press},\ \bibinfo {address} {Cambridge, MA, USA},\ \bibinfo {year} {2015})\
  pp.\ \bibinfo {pages} {2917--2925}\BibitemShut {NoStop}%
\bibitem [{\citenamefont {Mandt}\ \emph {et~al.}(2016)\citenamefont {Mandt},
  \citenamefont {Hoffman},\ and\ \citenamefont {Blei}}]{mandt16}%
  \BibitemOpen
  \bibfield  {author} {\bibinfo {author} {\bibfnamefont {S.}~\bibnamefont
  {Mandt}}, \bibinfo {author} {\bibfnamefont {M.~D.}\ \bibnamefont {Hoffman}},
  \ and\ \bibinfo {author} {\bibfnamefont {D.~M.}\ \bibnamefont {Blei}},\ }in\
  \href@noop {} {\emph {\bibinfo {booktitle} {Proceedings of the 33rd
  International Conference on International Conference on Machine Learning -
  Volume 48}}},\ \bibinfo {series and number} {ICML'16}\ (\bibinfo  {publisher}
  {JMLR.org},\ \bibinfo {year} {2016})\ pp.\ \bibinfo {pages}
  {354--363}\BibitemShut {NoStop}%
\bibitem [{\citenamefont {Chaudhari}\ and\ \citenamefont
  {Soatto}(2018)}]{chaudhari18}%
  \BibitemOpen
  \bibfield  {author} {\bibinfo {author} {\bibfnamefont {P.}~\bibnamefont
  {Chaudhari}}\ and\ \bibinfo {author} {\bibfnamefont {S.}~\bibnamefont
  {Soatto}},\ }in\ \href {\doibase 10.1109/ITA.2018.8503224} {\emph {\bibinfo
  {booktitle} {2018 Information Theory and Applications Workshop (ITA)}}}\
  (\bibinfo {year} {2018})\ pp.\ \bibinfo {pages} {1--10}\BibitemShut {NoStop}%
\bibitem [{\citenamefont {Chaudhari}\ \emph {et~al.}(2019)\citenamefont
  {Chaudhari}, \citenamefont {Choromanska}, \citenamefont {Soatto},
  \citenamefont {LeCun}, \citenamefont {Baldassi}, \citenamefont {Borgs},
  \citenamefont {Chayes}, \citenamefont {Sagun},\ and\ \citenamefont
  {Zecchina}}]{baldassi19}%
  \BibitemOpen
  \bibfield  {author} {\bibinfo {author} {\bibfnamefont {P.}~\bibnamefont
  {Chaudhari}}, \bibinfo {author} {\bibfnamefont {A.}~\bibnamefont
  {Choromanska}}, \bibinfo {author} {\bibfnamefont {S.}~\bibnamefont {Soatto}},
  \bibinfo {author} {\bibfnamefont {Y.}~\bibnamefont {LeCun}}, \bibinfo
  {author} {\bibfnamefont {C.}~\bibnamefont {Baldassi}}, \bibinfo {author}
  {\bibfnamefont {C.}~\bibnamefont {Borgs}}, \bibinfo {author} {\bibfnamefont
  {J.}~\bibnamefont {Chayes}}, \bibinfo {author} {\bibfnamefont
  {L.}~\bibnamefont {Sagun}}, \ and\ \bibinfo {author} {\bibfnamefont
  {R.}~\bibnamefont {Zecchina}},\ }\href {\doibase 10.1088/1742-5468/ab39d9}
  {\bibfield  {journal} {\bibinfo  {journal} {Journal of Statistical Mechanics:
  Theory and Experiment}\ }\textbf {\bibinfo {volume} {2019}},\ \bibinfo
  {pages} {124018} (\bibinfo {year} {2019})}\BibitemShut {NoStop}%
\bibitem [{\citenamefont {Saxe}\ \emph {et~al.}(2014)\citenamefont {Saxe},
  \citenamefont {McClelland},\ and\ \citenamefont {Ganguli}}]{ganguli13}%
  \BibitemOpen
  \bibfield  {author} {\bibinfo {author} {\bibfnamefont {A.~M.}\ \bibnamefont
  {Saxe}}, \bibinfo {author} {\bibfnamefont {J.~L.}\ \bibnamefont
  {McClelland}}, \ and\ \bibinfo {author} {\bibfnamefont {S.}~\bibnamefont
  {Ganguli}},\ }\href@noop {} {\enquote {\bibinfo {title} {Exact solutions to
  the nonlinear dynamics of learning in deep linear neural networks},}\ }
  (\bibinfo {year} {2014}),\ \Eprint {http://arxiv.org/abs/1312.6120v2}
  {arXiv:1312.6120v2 [cs.NE]} \BibitemShut {NoStop}%
\bibitem [{\citenamefont {Saxe}\ \emph {et~al.}(2019)\citenamefont {Saxe},
  \citenamefont {McClelland},\ and\ \citenamefont {Ganguli}}]{ganguli19}%
  \BibitemOpen
  \bibfield  {author} {\bibinfo {author} {\bibfnamefont {A.~M.}\ \bibnamefont
  {Saxe}}, \bibinfo {author} {\bibfnamefont {J.~L.}\ \bibnamefont
  {McClelland}}, \ and\ \bibinfo {author} {\bibfnamefont {S.}~\bibnamefont
  {Ganguli}},\ }\href {\doibase 10.1073/pnas.1820226116} {\bibfield  {journal}
  {\bibinfo  {journal} {Proceedings of the National Academy of Sciences}\
  }\textbf {\bibinfo {volume} {116}},\ \bibinfo {pages} {11537} (\bibinfo
  {year} {2019})},\ \Eprint
  {http://arxiv.org/abs/https://www.pnas.org/content/116/23/11537.full.pdf}
  {https://www.pnas.org/content/116/23/11537.full.pdf} \BibitemShut {NoStop}%
\bibitem [{\citenamefont {Bahri}\ \emph {et~al.}(2020)\citenamefont {Bahri},
  \citenamefont {Kadmon}, \citenamefont {Pennington}, \citenamefont
  {Schoenholz}, \citenamefont {Sohl-Dickstein},\ and\ \citenamefont
  {Ganguli}}]{ganguli20}%
  \BibitemOpen
  \bibfield  {author} {\bibinfo {author} {\bibfnamefont {Y.}~\bibnamefont
  {Bahri}}, \bibinfo {author} {\bibfnamefont {J.}~\bibnamefont {Kadmon}},
  \bibinfo {author} {\bibfnamefont {J.}~\bibnamefont {Pennington}}, \bibinfo
  {author} {\bibfnamefont {S.~S.}\ \bibnamefont {Schoenholz}}, \bibinfo
  {author} {\bibfnamefont {J.}~\bibnamefont {Sohl-Dickstein}}, \ and\ \bibinfo
  {author} {\bibfnamefont {S.}~\bibnamefont {Ganguli}},\ }\href {\doibase
  10.1146/annurev-conmatphys-031119-050745} {\bibfield  {journal} {\bibinfo
  {journal} {Annual Review of Condensed Matter Physics}\ }\textbf {\bibinfo
  {volume} {11}},\ \bibinfo {pages} {501} (\bibinfo {year} {2020})},\ \Eprint
  {http://arxiv.org/abs/https://doi.org/10.1146/annurev-conmatphys-031119-050745}
  {https://doi.org/10.1146/annurev-conmatphys-031119-050745} \BibitemShut
  {NoStop}%
\bibitem [{\citenamefont {Yaida}(2018)}]{yaida18}%
  \BibitemOpen
  \bibfield  {author} {\bibinfo {author} {\bibfnamefont {S.}~\bibnamefont
  {Yaida}},\ }\href@noop {} {\enquote {\bibinfo {title}
  {Fluctuation-dissipation relations for stochastic gradient descent},}\ }
  (\bibinfo {year} {2018}),\ \Eprint {http://arxiv.org/abs/1810.00004}
  {arXiv:1810.00004 [stat.ML]} \BibitemShut {NoStop}%
\bibitem [{\citenamefont {Ao}(2004)}]{ao04}%
  \BibitemOpen
  \bibfield  {author} {\bibinfo {author} {\bibfnamefont {P.}~\bibnamefont
  {Ao}},\ }\href {\doibase 10.1088/0305-4470/37/3/l01} {\bibfield  {journal}
  {\bibinfo  {journal} {Journal of Physics A: Mathematical and General}\
  }\textbf {\bibinfo {volume} {37}},\ \bibinfo {pages} {L25} (\bibinfo {year}
  {2004})}\BibitemShut {NoStop}%
\bibitem [{\citenamefont {Kwon}\ \emph {et~al.}(2005)\citenamefont {Kwon},
  \citenamefont {Ao},\ and\ \citenamefont {Thouless}}]{KAT05}%
  \BibitemOpen
  \bibfield  {author} {\bibinfo {author} {\bibfnamefont {C.}~\bibnamefont
  {Kwon}}, \bibinfo {author} {\bibfnamefont {P.}~\bibnamefont {Ao}}, \ and\
  \bibinfo {author} {\bibfnamefont {D.~J.}\ \bibnamefont {Thouless}},\ }\href
  {\doibase 10.1073/pnas.0506347102} {\bibfield  {journal} {\bibinfo  {journal}
  {Proceedings of the National Academy of Sciences}\ }\textbf {\bibinfo
  {volume} {102}},\ \bibinfo {pages} {13029} (\bibinfo {year} {2005})},\
  \Eprint
  {http://arxiv.org/abs/https://www.pnas.org/content/102/37/13029.full.pdf}
  {https://www.pnas.org/content/102/37/13029.full.pdf} \BibitemShut {NoStop}%
\bibitem [{\citenamefont {Yin}\ and\ \citenamefont {Ao}(2006)}]{ao06}%
  \BibitemOpen
  \bibfield  {author} {\bibinfo {author} {\bibfnamefont {L.}~\bibnamefont
  {Yin}}\ and\ \bibinfo {author} {\bibfnamefont {P.}~\bibnamefont {Ao}},\
  }\href {\doibase 10.1088/0305-4470/39/27/003} {\bibfield  {journal} {\bibinfo
   {journal} {Journal of Physics A: Mathematical and General}\ }\textbf
  {\bibinfo {volume} {39}},\ \bibinfo {pages} {8593} (\bibinfo {year}
  {2006})}\BibitemShut {NoStop}%
\bibitem [{\citenamefont {Kwon}\ \emph {et~al.}(2011)\citenamefont {Kwon},
  \citenamefont {Noh},\ and\ \citenamefont {Park}}]{kwon11a}%
  \BibitemOpen
  \bibfield  {author} {\bibinfo {author} {\bibfnamefont {C.}~\bibnamefont
  {Kwon}}, \bibinfo {author} {\bibfnamefont {J.~D.}\ \bibnamefont {Noh}}, \
  and\ \bibinfo {author} {\bibfnamefont {H.}~\bibnamefont {Park}},\ }\href
  {\doibase 10.1103/PhysRevE.83.061145} {\bibfield  {journal} {\bibinfo
  {journal} {Phys. Rev. E}\ }\textbf {\bibinfo {volume} {83}},\ \bibinfo
  {pages} {061145} (\bibinfo {year} {2011})}\BibitemShut {NoStop}%
\bibitem [{\citenamefont {Kwon}\ and\ \citenamefont {Ao}(2011)}]{kwon11}%
  \BibitemOpen
  \bibfield  {author} {\bibinfo {author} {\bibfnamefont {C.}~\bibnamefont
  {Kwon}}\ and\ \bibinfo {author} {\bibfnamefont {P.}~\bibnamefont {Ao}},\
  }\href {\doibase 10.1103/PhysRevE.84.061106} {\bibfield  {journal} {\bibinfo
  {journal} {Phys. Rev. E}\ }\textbf {\bibinfo {volume} {84}},\ \bibinfo
  {pages} {061106} (\bibinfo {year} {2011})}\BibitemShut {NoStop}%
\bibitem [{\citenamefont {Risken}\ and\ \citenamefont
  {Frank}(1996)}]{risken96}%
  \BibitemOpen
  \bibfield  {author} {\bibinfo {author} {\bibfnamefont {H.}~\bibnamefont
  {Risken}}\ and\ \bibinfo {author} {\bibfnamefont {T.}~\bibnamefont {Frank}},\
  }\href@noop {} {\emph {\bibinfo {title} {The Fokker-Planck Equation}}}\
  (\bibinfo  {publisher} {Springer},\ \bibinfo {address} {Berlin, Heidelberg},\
  \bibinfo {year} {1996})\BibitemShut {NoStop}%
\bibitem [{\citenamefont {G.~Cohen}\ and\ \citenamefont {van
  Schaik}(2017)}]{cohen17}%
  \BibitemOpen
  \bibfield  {author} {\bibinfo {author} {\bibfnamefont {J.~T.}\ \bibnamefont
  {G.~Cohen}, \bibfnamefont {S.~Afshar}}\ and\ \bibinfo {author} {\bibfnamefont
  {A.}~\bibnamefont {van Schaik}},\ }\href@noop {} {\enquote {\bibinfo {title}
  {Emnist: an extension of mnist to handwritten letters},}\ } (\bibinfo {year}
  {2017}),\ \Eprint {http://arxiv.org/abs/1702.05373} {arXiv:1702.05373
  [cs.NE]} \BibitemShut {NoStop}%
\end{thebibliography}%
\end{document}